\documentclass[sn-apa,iicol]{sn-jnl}% Default with double column layout
\usepackage{amsmath}
\usepackage{amssymb}
\usepackage{graphicx}

\usepackage{booktabs}
\usepackage{multirow}
\usepackage{bbm}

\jyear{2021}%

\theoremstyle{thmstyleone}%
\theoremstyle{thmstyletwo}%

\theoremstyle{thmstylethree}%

\begin{document}

\title[RAC-Net]{Reliability-Adaptive Consistency Regularization for Weakly-Supervised Point Cloud Segmentation}


\author[1]{\fnm{Zhonghua} \sur{Wu}}\email{zhonghua001@e.ntu.edu.sg}
\equalcont{These authors contributed equally to this work.}

\author[2]{\fnm{Yicheng} \sur{Wu}}\email{yicheng.wu@monash.edu}
\equalcont{These authors contributed equally to this work.}

\author*[1]{\fnm{Guosheng} \sur{Lin}}\email{gslin@ntu.edu.sg}

\author[2]{\fnm{Jianfei} \sur{Cai}}\email{jianfei.cai@monash.edu}

\affil[1]{\orgdiv{School of Computer Science and Engineering}, \orgname{Nanyang Technological University}, \orgaddress{\city{Singapore}, \postcode{639798}, \country{Singapore}}}

\affil[2]{\orgdiv{Department of Data Science and AI}, \orgname{Monash University}, \orgaddress{\city{Melbourne}, \postcode{VIC 3800}, \country{Australia}}}


\abstract{Weakly-supervised point cloud segmentation with extremely limited labels is highly desirable to alleviate the expensive costs of collecting densely annotated 3D points. This paper explores applying the consistency regularization that is commonly used in weakly-supervised learning, for its point cloud counterpart with multiple data-specific augmentations, which has not been well studied. We observe that the straightforward way of applying consistency constraints to weakly-supervised point cloud segmentation has two major limitations: noisy pseudo labels due to the conventional confidence-based selection and insufficient consistency constraints due to discarding unreliable pseudo labels. Therefore, we propose a novel \textbf{R}eliability-\textbf{A}daptive \textbf{C}onsistency Network (RAC-Net) to use both prediction confidence and model uncertainty to measure the reliability of pseudo labels and apply consistency training on all unlabeled points while with different consistency constraints for different points based on the reliability of corresponding pseudo labels. Experimental results on the S3DIS and ScanNet-v2 benchmark datasets show that our model achieves superior performance in weakly-supervised point cloud segmentation. The code will be released publicly at \url{https://github.com/wu-zhonghua/RAC-Net}.}




\maketitle

\section{Introduction}
\label{sec:intro}
Recently, 3D point cloud segmentation has achieved impressive progresses~\citep{choy20194d, thomas2019kpconv, qi2017pointnet, qi2017pointnet++,hu2022sensaturban}. However, it is still extremely expensive and labor-consuming to collect abundant point-level annotations for the model training. Therefore, in order to alleviate huge labeling costs, it is highly desirable to develop weakly-supervised point cloud segmentation, which aims to train a satisfied segmentation model with scarce labeled points but enormous unlabeled points. 

\begin{figure}[t]
\centering
    \includegraphics[width=\linewidth]{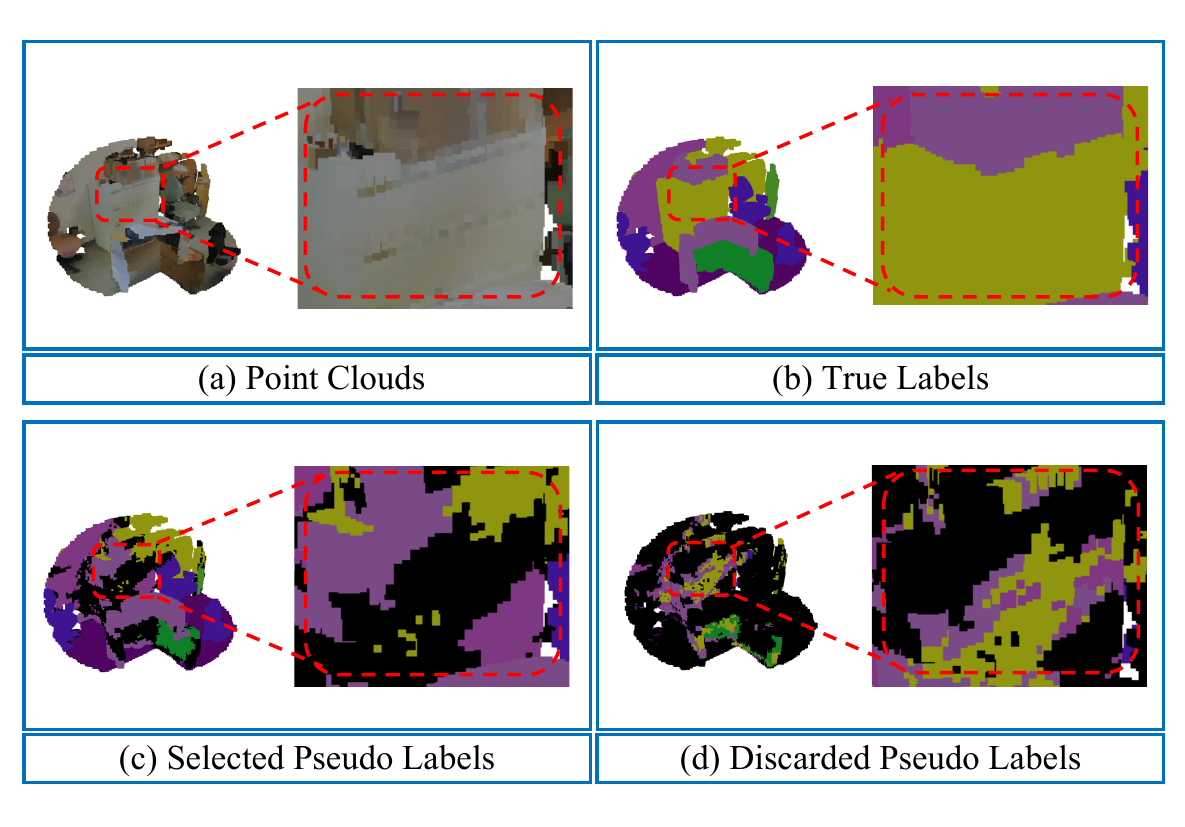}
    \caption{Illustrations of imperfect pseudo labels in the conventional confidence-based selection. We use a probability threshold of 0.7 to select highly-confident pseudo labels for the model training, while they are very noisy (b vs. c), and many discarded pseudo labels (d) are not exploited during training.}
    \label{fig:cb}
\end{figure}

To exploit the unlabeled points, existing methods are mainly based on the consistency assumption~\citep{zhang2021flexmatch, abuduweili2021adaptive, yuan2021simple,wu2022dual}, where the model is encouraged to be consistent under various perturbations, to achieve the local distributional smoothness (LDS).
For example, \citet{sohn2020fixmatch} utilized the predictions of weakly augmented data to guide the learning of strongly augmented versions, where they select reliable predictions as pseudo labels based on the prediction confidence and use them to enforce the consistency constraints to regularize the model training.
Such consistency-based regularization has not been well investigated for weakly-supervised point cloud segmentation. For instance, the recent 1T1C~\citep{liu2021one} model also leverages the confidence scores to select reliable predictions as pseudo labels and uses them to train the model iteratively, which however, is not a consistency constraint under diversified perturbations.

This motivates us to study the intuitive idea of applying consistency constraints to improve weakly-supervised point cloud segmentation. The straightforward way is to directly extend the FixMatch~\citep{sohn2020fixmatch} from images to point clouds, \textit{i.e.,} selecting confident predictions of the weakly augmented point clouds as pseudo labels and applying consistency constraints to guide the predictions of strongly augmented ones. However, such a scheme has two major limitations. First, it is unsatisfactory to select reliable predictions based on their confidence. The examples in Fig.~\ref{fig:cb} (b, c) illustrate that the scheme may generate highly confident but incorrect pseudo labels, which would lead to more noisy supervision and confuse the model training. Second, for the large amounts of unlabeled points that are deemed unreliable (see Fig.~\ref{fig:cb} (d)), they are being discarded and not utilized during training~\citep{sohn2020fixmatch,liu2021one}, resulting in sub-optimal performance.

We would like to point out \textit{these limitations are particularly noticeable for weakly-supervised point cloud segmentation, while they might not be so significant in the corresponding image counterpart}. This is because for weakly-supervised point cloud segmentation, the human annotations are much more scarce, \textit{e.g.} the typical one thing one click (OTOC) setting~\citep{liu2021one}, and the generated point cloud pseudo labels are much more noisy (due to the sparsity of point clouds and lack of neighbor support). Thus, the key questions for weakly-supervised point cloud segmentation are: \textit{how to select reliable pseudo labels and how to utilize a large number of unreliable pseudo labels?}

Our key idea in this work is to 
\textit{select more reliable pseudo labels by considering both prediction confidence and model uncertainty, and utilize reliable predictions as hard pseudo labels while using ambiguous predictions as soft pseudo labels instead of throwing them away}. Specifically, we propose a simple yet effective \textit{\textbf{R}eliability-\textbf{A}daptive \textbf{C}onsistency Network} (RAC-Net), which enforces the consistency constraints on all unlabeled data adaptively based on their pseudo label reliability. To measure the reliability, we jointly use the prediction confidence and uncertainty to divide the initial predictions of unlabeled data into ambiguous and reliable sets, where the uncertainty is measured by computing the statistical variances among the predictions of different augmentations. Considering the ambiguous predictions are unreliable, we treat them as soft pseudo labels and apply a consistency loss (KL Divergence) to encourage invariant results of augmented point clouds. Considering the reliable predictions are accurate, we convert them into one-hot pseudo labels and then apply a consistency loss (Cross-entropy Loss) to guide the learning of different augmented data. In addition, to further exploit the reliable set, we also generate mix-augmented point clouds by a point-wise interpolation among multiple off-the-shelf base-augmentations and then use the one-hot pseudo labels to facilitate the model training. 

We follow the public models~\citep{liu2021one, wu2022dual} to conduct experiments on two large-scale point cloud segmentation datasets: S3DIS \citep{armeni20163d} and ScanNet-v2  \citep{dai2017scannet}. Extensive experiments demonstrate that our RAC-Net is able to accurately select pseudo labels during the model training and achieves superior segmentation performance than other existing methods for weakly-supervised point cloud segmentation, \textit{e.g., outperforming the DAT model \citep{wu2022dual} by a 1.9\% mIoU gain under the OTOC setting on the S3DIS dataset}. Besides, our experimental results reveal that combining the local shape deformation like PointWolf~\citep{kim2021point} and the conventional augmentation (\textit{e.g.,} Affine Transformations) is able to achieve impressive performance gains for weakly-supervised point cloud segmentation.

Overall, the contributions of this paper can be summarized as follows:
\begin{itemize}
\item  
We consider the problem of applying consistency-based regularization for weakly-supervised point cloud segmentation and identify the two main obstacles: measuring pseudo-label reliability and utilizing unreliable pseudo-labels.
\item We propose a novel RAC-Net to incorporate both the prediction confidence and uncertainty, which is computed as the discrepancy among different augmentations to identify more reliable pseudo-labels, and adaptively apply different consistency constraints to different points based on their reliability. Moreover, we design a mix-augmentation module to generate mix-augmented point clouds to further exploit the high-quality reliable set.
\item We investigate various point cloud-specific augmentation strategies (\textit{e.g.,} PointWolf) and carry out comprehensive experiments to demonstrate the effectiveness of each augmentation.
\item We introduce an interpolation strategy that leverages both local and global spatial transformations to generate strongly augmented samples, which is a complement of strong augmentations in point cloud tasks.
\item  Our proposed RAC-Net achieves new state-of-the-art performance in weakly-supervised point cloud segmentation on S3DIS and ScanNet-v2 datasets, with multiple base augmentations and mix-augmentations for adaptive consistency training. 
\end{itemize}

\begin{figure*}[t]
\centering
\includegraphics[width=\linewidth]{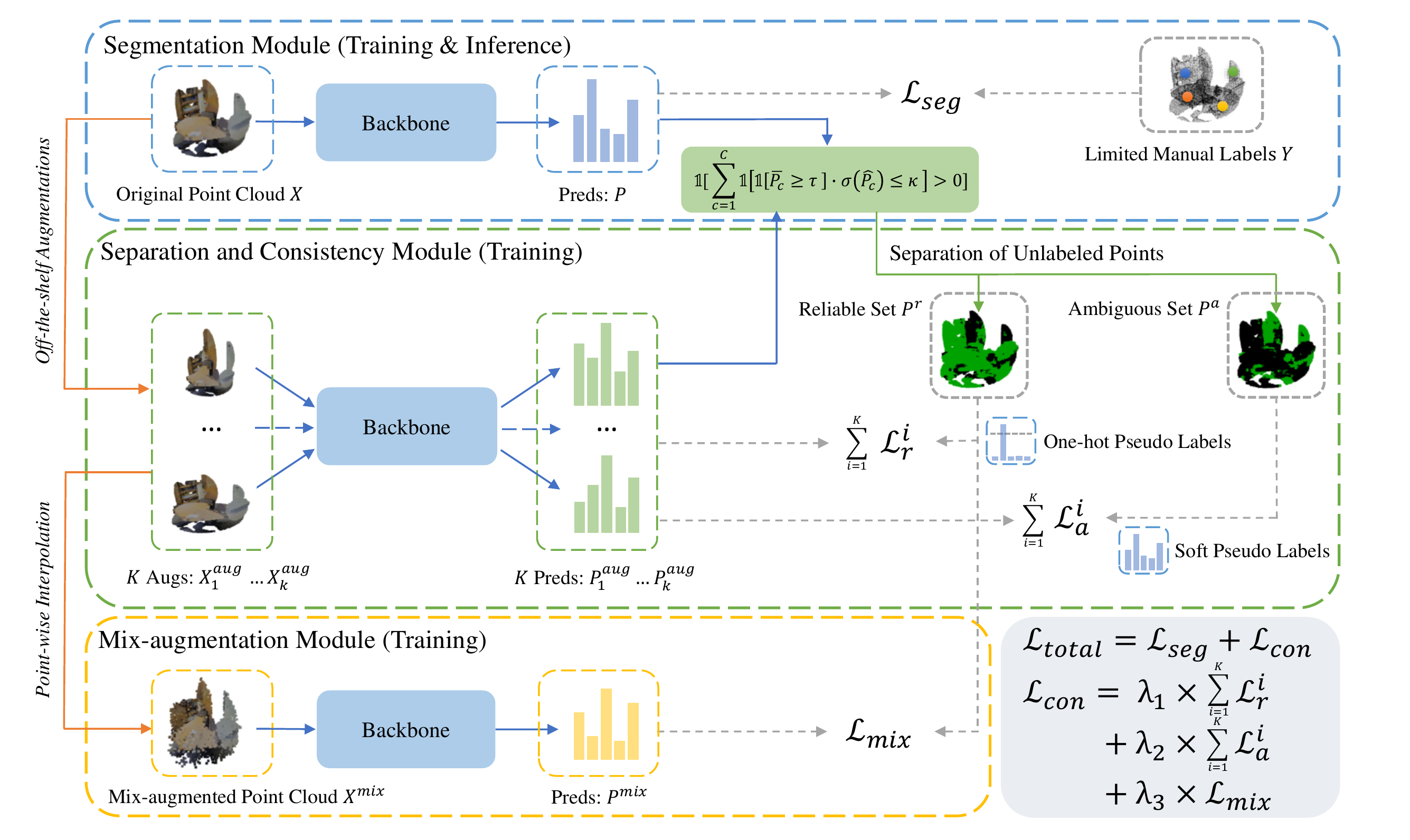}
    \caption{An overview of our RAC-Net, which includes a Segmentation Module (Top) to utilize limited sparse annotations, a Separation and Consistency Module (Middle) to separate the unlabeled points into reliable and ambiguous sets according to their reliability (prediction confidence and uncertainty) and apply different consistency constraints for different sets, and a Mix-augmentation Module (Bottom) to sufficiently exploit the one-hot reliable pseudo labels with mix-augmentations. Note that, only the segmentation module is included at the inference stage. Therefore, there is no extra inference cost.}
    \label{fig:pipeline}
\end{figure*}

\section{Related Works}
\subsection{Weakly-supervised Point Cloud Segmentation}
Many works \citep{cheng2021sspc,zhao2021few, wang2022new,ye2022efficient,gongoptimization,pang2022masked,cheraghian2022zero} aim to reduce the huge human labeling costs of annotating 3D points. For example, \citet{wei2020multi} introduced a sub-cloud level annotation scheme that first divides a point cloud into a few sub-clouds and then only annotates the class labels that appear in each sub-cloud. With the sub-cloud level annotations, they generated the pseudo segmentation masks via the Class Activation Maps and then used them for training. However, their segmentation performance is unsatisfactory due to the lack of localization information. To address the issue, \citet{xu2020weakly} proposed to label 10\% points of one entire point cloud and introduced three losses from self-supervision, inexact-supervision, and point smoothness perspectives to leverage the unlabeled data. In order to further reduce the number of labeled points and preserve the localization information, \citet{liu2021one} introduced the OTOC annotation scheme that only labels one point for each thing in the whole scene. With the sparse annotations, they further adopted a confidence-based method to select pseudo labels to train the model iteratively. However, this selection scheme introduces many inaccurate pseudo labels, resulting in sub-optimal performance. The PSD~\citep{zhang2021perturbed} method is a perturbed self-distillation framework to ensure the predictive consistency on all the points among original samples and perturbed ones. Meanwhile, \citet{wu2022dual} employed adversarial training to enforce a stronger perturbation. PointMatch~\citep{wu2022pointmatch} applied multi-view regularization and used the super-points to generate the pseudo labels. Furthermore, \citet{Yang_2022_CVPR} proposed a transformer-based model to improve the weakly-supervised point cloud segmentation.

Recently, HybridCR \citep{li2022hybridcr} proposed to leverage both point consistency and contrastive properties for weakly-supervised point cloud semantic segmentation in an end-to-end manner. Specifically, it introduced local and global guidance contrastive regularization to enhance high-level 3D semantic scene understanding tasks. Additionally, it incorporated a novel dynamic point cloud augmentor to transform diverse and robust sample views, which are jointly optimized throughout the entire training process. Box2Mask \citep{chibane2022box2mask} focuses on weakly supervised instance segmentation tasks with box-level supervision. The method was designed to utilize bounding boxes both as a representation and as a guide for the training scheme. The Back to Reality method \citep{xu2022back} was proposed to make use of synthetic 3D shapes, converting weak labels into fully-annotated virtual scenes to provide stronger supervision. It then leveraged these perfect virtual labels to complement and refine the real labels. Shi et al. \citep{shi2022weakly} extended temporal matching and spatial graph propagation techniques for weakly supervised 4D point cloud segmentation tasks.

In this research, our proposed RAC-Net does not rely on pre-defined super-points or particular backbones and adaptively applies different consistency regularizations to train reliable and ambiguous points. Our simple adaptive consistency regularization strategy achieves impressive performance improvements and it can be easily integrated with other existing methods \citep{wu2022dual, Yang_2022_CVPR}.

\subsection{Consistency Regularization}
Weakly supervised learning~\citep{liu2023harmonizing, wu2021learning, wu2020exploring, wu2019keypoint} attracts much attention since it can reduce the heavy burden of collecting well-annotated data. A widely used approach in semi-supervised learning is based on consistency regularization, aiming to achieve a local distribution smoothness (LDS) with certain perturbations. The consistency constraint~\citep{zhong2021pixel,alonso2021semi,saito2021openmatch,wu2022exploring,fan2022revisiting,zang2023semi,lopez2022desc, wu2023coactseg} is widely used to leverage unlabeled data. For example, \citet{ouali2020semi} applied both the data-level and feature-level perturbations to perturb unlabeled data. \citet{wu2022mutual} proposed to encourage the invariant results generated by different decoders to leverage the unlabeled data, as a model-level perturbation way. MixMatch~\citep{berthelot2019mixmatch} used the mix-up operation to combine different samples to enrich the contexts for the model training. The VAT model \citep{miyato2018virtual} studied the adversarial training for the consistency regularization to leverage the unlabeled points, as a stronger smoothness constraint.  

Pseudo-label learning is also a type of consistency training way. It usually utilizes the confidence-based strategy \citep{sohn2020fixmatch, rizve2021defense}, which ensures that unlabeled data is utilized only when the model's predictions are highly confident. However, there is a challenge with this method as the selection of unlabeled samples based on high-confidence predictions can move decision boundaries to low-density regions \citep{chapelle2005semi}, resulting in the inclusion of many incorrect predictions. This issue arises from the poor calibration of neural networks \citep{guo2017calibration}, which refers to the discrepancy between a network's individual prediction confidences and its overall accuracy.

In this paper, we also follow the consistency training pipeline and introduce customized techniques to perturb 3D points sufficiently. Most importantly, our consistency training is adaptive to the reliability of pseudo labels. At the same time, to overcome the issue of selecting wrong pseudo labels, we here consider both prediction confidence and model uncertainty to select the most reliable pseudo labels.

\subsection{Noise Learning}

Several techniques \citep{xiao2015learning, goldberger2016training, bekker2016training, liu2024lcreg} have been proposed for effectively training accurate models under conditions of noisy labeled data. The noise-robust layers-based methods are designed to predict a label transition matrix $T$. By using the estimated matrix $T$, these methods can adjust the output of the network to a more confident label. However, they mainly rely on a strong correlation between labels, thus restricting their applicability. Other methods try to design a loss function that remains robust in the face of noisy labels. For example, Generalized Cross Entropy (GCE) \citep{zhang2018generalized} and Symmetric Cross Entropy (SCE) \citep{wang2019symmetric} can be easily adapted to existing architectures. However, one of the limitations of these methods is that they are hard to handle severely noisy labels. Loss adjustment techniques aim to diminish the adverse effects of noisy labels by altering the loss across all training samples. While these methods leverage all available training data, they run the risk of false correction. To avoid false correction, sample selection methods \citep{shen2019learning, song2019selfie} circumvent false corrections by selecting accurately labeled samples from noisy data. These methods are mainly designed for image-based classification and segmentation tasks. For the point cloud segmentation tasks, Ye et.al \citep{ye2021learning} proposed a hybrid learning scheme including sample selection and loss correction to learn a robust model with noisy labels.

In this work, unlike previous methods only consider prediction confidence to select the labels, we further consider model uncertainty to select reliable pseudo labels, which can be regarded as the hard pseudo labels for the model training. Moreover, we utilize ambiguous predictions as soft pseudo labels instead of throwing them away to further boost the performance.

\subsection{Uncertainty Estimation}
Approximating the prediction uncertainty~\citep{graves2011practical,malinin2018predictive, uncertainty,chen2022hyperbolic, alter1998uncertainty} of the deep model has been widely studied in computer vision. For example, \citet{yu2019uncertainty} proposed to leverage MC-Dropout to estimate uncertainty and used it to improve the consistency learning for medical image segmentation. UPS~\citep{rizve2021defense} employed uncertainty for the selection of pseudo labels. Similarly, \citet{mukherjee2020uncertainty} used the uncertainty to select pseudo labels from a pre-trained language model for the downstream semi-supervised tasks. 

Here, the uncertainty is measured based on the prediction discrepancy among different augmentations, which is combined with the confidence for the reliability measurement of unlabeled points.

\section{Methods}
As illustrated in Figure~\ref{fig:pipeline}, our RAC-Net consists of three parts for weakly-supervised point cloud segmentation: (1) A segmentation module is used to train the model with limited sparse annotations. (2) A separation and consistency module considers both prediction confidence and uncertainty to divide the unlabeled points into two sets: reliable and ambiguous ones. Then the consistency constraints are applied with one-hot and soft pseudo labels on the reliable and ambiguous sets, respectively. (3) A mix-augmentation module further enforces the consistency constraints for the reliable points with a mix-augmented technique, to sufficiently exploit these high-quality pseudo labels.

\subsection{Segmentation Module}
Consider the following notations. The input set is denoted as $X = [L, F] \in \mathbb{R}^{N \times (3+D_f)}$, which includes $N$ points containing the point locations $L \in \mathbb{R}^{N \times 3}$ and the corresponding features $F \in \mathbb{R}^{N \times D_f}$. We use $Y\in \mathbb{R}^{M \times 1}$ to denote limited manual labels, where only $M$ points have their corresponding true labels ($M << N$). With the segmentation model $f(\theta)$, its prediction of the $i$-$th$ point $x_i$ is denoted as $p(\hat{y_i} \rvert x_i;\theta) \in P$, $i \in \{1, ..., N\}$. During training, we apply a cross-entropy (CE) loss $\mathcal{L}_{seg}$ to supervise our model with the guidance of the limited labels $Y$.

\subsection{Separation and Consistency Module}
To exploit the unlabeled points, we first divide them into reliable and ambiguous sets. Normally, the reliability is measured by the confidence scores of its prediction~\citep{liu2021one}, where the sample is regarded as reliable if the confidence score exceeds a threshold. However, such a strategy often leads to incorrect pseudo labels~\citep{arazo2020pseudo}. Specifically, the model might generate highly confident but wrong predictions, which confuse the model training. To address this issue, we propose to further incorporate the uncertainty into the measurement of reliability, aiming at accurately dividing the pseudo labels. In contrast to the conventional uncertainty measurement methods~\citep{rizve2021defense}, in the point cloud segmentation tasks, invariance under transformations is important for the model to capture the features of 3D objects. Thus, we propose to use the prediction discrepancy among different augmentations to measure the model uncertainty.

Specifically, as shown in Figure~\ref{fig:pipeline}, we first generate $K$ augmented point clouds $X^{aug}_{1} ... X^{aug}_{k}$ of the original point cloud $X$ with multiple off-the-shelf augmentation methods (\textit{e.g.,} PointWolf~\citep{kim2021point}). Then, we generate the predictions for them (labeled as $P^{aug}_{1}, ..., P^{aug}_{k}$) and obtain a prediction set $\hat{P}$ containing $K$ predictions and the original predicted results $P$. Afterwards, we compute the statistical variance as the uncertainty $\sigma(\hat{P})$ and obtain the confidence as the mean of the $K+1$ predictions, denoted as $\bar{P}$.

\begin{equation}
  \bar{P} = \frac{P + \sum_{i=1}^{K}P_i^{aug}}{K+1}
\end{equation}
\begin{equation}
  \sigma(\hat{P}) = \sqrt{\frac{(P - \bar{P})^2 + \sum_{i=1}^{K}(P_i^{aug} - \bar{P})^2}{K + 1} }
\end{equation}
Then, we use both the confidence and uncertainty to divide the pseudo labels $P$ into the reliable set $P^{r}$ and the ambiguous set $P^{a}$:
\begin{equation}
\begin{aligned}
& P^{r} =  R \cdot P , P^{a} = (1 - R) \cdot P, \\
& R = \mathbbm{1}[\sum^C_{c=1} (\mathbbm{1}[\bar{P_c} \geq \tau ] \cdot \mathbbm{1}[\sigma(\hat{P}_c) \leq \kappa ])  > 0], \\
\end{aligned}
\label{eq:division}
\end{equation}
where $\tau$ and $\kappa$ are two pre-defined thresholds corresponding to confidence or uncertainty respectively, 
$C$ denotes the number of classes, and $\mathbbm{1}$ is the indicator function. Essentially, the binary mask $R$ will select the predictions to the reliable set $P^{r}$ if their values in one class are consistently high confident across different augmentations. Conversely, the remaining predictions with low confidence or high uncertainty among different augmented versions are regarded as the ambiguous predictions $P^{a}$.

For the reliable predictions $P^r$, considering they are accurate, we first convert them to the one-hot pseudo labels $\widetilde{Y}$ via an $argmax$ operation. Then we enforce another consistency constraint by applying a cross-entropy loss on the predictions of various augmentations: 
\begin{equation}
\begin{aligned}
&\mathcal{L}_{r} = \sum_{i=1}^{K}\mathrm{CE}\left[\widetilde{Y},  R \cdot P_i^{aug}\right].
\end{aligned}
\end{equation}

Considering the ambiguous predictions $P^{a}$ are with high uncertainty or low confidence, we treat them as soft pseudo labels, and only apply a consistency constraint on augmented data to boost the model training. Specifically, we use KL Divergence between the soft pseudo labels $P^{a}$ and the predictions of all augmented versions as
\begin{equation}
\begin{aligned}
\mathcal{L}_{a} = \sum_{i=1}^{K}\mathrm{KL}\left[P^a, (1 - R) \cdot P_i^{aug} \right].
\end{aligned}
\end{equation}

\subsection{Mix-augmentation Module}

Moreover, as Fig.~\ref{fig:pipeline} shows, we further generate the mix-augmented point clouds and use the one-hot reliable pseudo labels for the model training. Here, our interpolated strategy can produce strongly augmented samples by applying both the local and global spatial transformations. In this way, the reliable pseudo labels are sufficiently leveraged to guide the training in a typical weak-strong learning scheme \citep{sohn2020fixmatch}. Particularly, we first randomly select two base-augmented point clouds $X^{aug}_{m}$ and $X^{aug}_{n}$ from the $K$ augmentations. Then we combine them via a point-wise interpolation operation to generate mix-augmented data $X^{mix}$ as 
\begin{equation}
X^{mix} = \alpha \cdot X^{aug}_{m} + (1 - \alpha) \cdot X^{aug}_{n}.
\end{equation}
where $\alpha \in \mathbb{R}^{N \times 1}$ is a sampling probability following the uniform distribution. Note that, when $K=1$, we generate the mix-augmented point cloud $X^{mix}$ by the point-wise interpolation operation between the original point cloud and its augmented version. We then obtain the predictions $P^{mix}$ of $X^{mix}$. Finally, we adopt the CE loss to supervise $P^{mix}$ with the reliable pseudo-labels $\widetilde{Y}$:
\begin{equation}
\begin{aligned}
&\mathcal{L}_{mix} = \mathrm{CE}\left[\widetilde{Y},  R \cdot P^{mix}\right].
\end{aligned}
\end{equation}

To achieve a trade-off between the effectiveness and efficiency, we set $K$ as 2 and employ two popular point cloud augmentation methods, \textit{i.e.,} PointWolf and Affine Transformations. We refer the readers to Section~\ref{sec:ab_K} for more discussions about them. Note that, other augmentation techniques \citep{wu2022dual,wu2022pointmatch,miyato2018virtual} also can be incorporated into our framework to further improve the performance.

Finally, the total loss of our RAC-Net is a weight sum of $\mathcal{L}_{seg}$, $\mathcal{L}_{r}$, $\mathcal{L}_{a}$ and $\mathcal{L}_{mix}$:
\begin{equation}
\mathcal{L}_{total} =\mathcal{L}_{seg} + \lambda_1\mathcal{L}_{r} + \lambda_2 \mathcal{L}_{a} + \lambda_3\mathcal{L}_{mix},
\end{equation}
where the weights $\lambda_1$, $\lambda_2$, $\lambda_3$ are set as 1 in our experiments for simplicity. Note that, $\mathcal{L}_{seg}$ is only used for the limited annotated points, and other losses are used to regularize the learning of all data. 

\section{Experiments}
\subsection{Implementation Details}

\noindent\textbf{Datasets.} 
\label{sec:data}
We follow the 1T1C, DAT and SQN models~\citep{liu2021one,wu2022dual,hu2021sqn} to conduct experiments on the S3DIS~\citep{armeni20163d}, ScanNet-v2~\citep{dai2017scannet} datasets and SemanticKitti dataset \citep{behley2019semantickitti}, for a fair comparison. The S3DIS dataset consists of 3D scans of 271 rooms belonging to 6 areas with 13 categories. We train the segmentation model on Area 1, 2, 3, 4, 6 and test it on Area 5, respectively. The ScanNet-v2 dataset contains 1513 3D scans with 20 categories, which are divided into 1201, 312, and 100 scans for training, validation, and testing, respectively.

\begin{table}
% \vspace{-2cm}
\caption{Comparison of our model and several public methods on the S3DIS testing set. The proposed RAC-Net achieves superior performance under both the OTOC and OTTC settings.}
\centering

\resizebox{\linewidth}{!}{
\begin{tabular}{c|c|c}
\toprule
Method                                                               & Supervision & mIoU (\%)                      \\ \midrule
PointNet \citep{qi2017pointnet}                 & 100\%            & 41.1                           \\
PointCNN \citep{li2018pointcnn}                 & 100\%            & 57.3                           \\
Xu et al. \citep{xu2020weakly}                  & 0.2\%            & 44.5                           \\
Xu et al. \citep{xu2020weakly}                  & 10\%             & 48.0                           \\
GPFN \citep{wang2020weakly}                     & 16.7\%+2D        & 50.8                           \\
GPFN \citep{wang2020weakly}                     & 100\%+2D         & 52.5                           \\ \midrule
Our Baseline (KPConv)                        & 0.02\%    & 50.1 \\
1T1C \citep{liu2021one}                         & 0.02\% (OTOC)    & 50.1                           \\
MIL-derived transformer \citep{Yang_2022_CVPR}                         & 0.02\% (OTOC)    & 51.4                           \\ 
DAT \citep{wu2022dual}                         & 0.02\% (OTOC)    & 56.5                           \\
PointMatch \citep{wu2022pointmatch}                         & 0.02\% (OTOC)    & 55.3                           \\ 
Our RAC-Net (KPConv)                                                           & 0.02\% (OTOC)    & \textbf{58.4} \\

\midrule
Our Baseline (KPConv)                        & 0.06\%    & 54.3 \\
1T1C \citep{liu2021one}                         & 0.06\% (OTTC)    & 55.3                           \\ 
DAT \citep{wu2022dual}                        & 0.06\% (OTTC)    & 58.5                           \\

Our RAC-Net (KPConv)                                                         & 0.06\% (OTTC)    & \textbf{60.5} \\ \midrule 
Our Upper Bound (KPConv)                                                  & 100\%            & 65.4                           \\ \bottomrule
\end{tabular}
}
\label{tab:s3dis-compare-sota}
\end{table}

\begin{table}

\caption{Ablation studies of our RAC-Net under the OTOC setting on the S3DIS dataset.}
\centering
\begin{tabular}{cccc|c}
\toprule
 \multicolumn{1}{c|}{$\mathcal{L}_{seg}$} & \multicolumn{1}{c|}{$\mathcal{L}_{a}$} & \multicolumn{1}{c|}{$\mathcal{L}_{r}$} & $\mathcal{L}_{mix}$ & mIoU(\%) \\ \midrule
    \checkmark                  &                                           &                                             &                     & 50.1 \\
     \checkmark                 &                                &                     \checkmark                         &                     &   51.0   \\
         \checkmark             &                                 &             
         &   \checkmark        & 52.1 \\
         \checkmark             &                                 &                     \checkmark                        &   \checkmark        & 52.5 \\
         \midrule
    \checkmark                  &        \checkmark                                   &                                   &                     &  56.3    \\
\checkmark                      &       \checkmark                          &               \checkmark                    &                     & 57.6 \\
         \checkmark             &            \checkmark                      &             
         &   \checkmark        & 57.0\\
         
    \checkmark                  &     \checkmark                            &            \checkmark                       &     \checkmark      & \textbf{58.4} \\ \bottomrule
\end{tabular}
\label{tab:ablation-loss}

\end{table}

\noindent\textbf{Weak Annotation Scheme.}
On the S3DIS dataset, we follow existing methods~\citep{liu2021one, wu2022dual} to annotate the training data under the ``OTOC'' setting. For each object, we randomly select a point as the labeled one with the same probability. Consequently, only 0.02\% of points have been labeled in the whole dataset. For the ScanNet-v2 dataset, we conduct experiments on the ``3D Semantic label with Limited Annotations'' benchmark~\citep{dai2017scannet}, where only 20 fixed points are labeled in each room scene. For the SemanticKitti dataset, we follow SQN~\citep{hu2021sqn} to conduct experiments under the ``OTOC'' setting, where only 0.01\% points have been labeled during the model training.

\begin{table*}
\caption{Discussion of our proposed RAC-Net under the OTOC setting on the S3DIS dataset. Note that our method sets the thresholds $\tau$ and $\kappa$ as 0.7 and 0.05, respectively, to employ both prediction confidence and uncertainty to divide the unlabeled points.
}
\centering
\resizebox{0.6\linewidth}{!}{

\begin{tabular}{c|cc|c}
\toprule
\multirow{2}{*}{Division Strategy} & \multicolumn{2}{c|}{Consistency Loss}               & \multirow{2}{*}{mIoU (\%)}     \\ \cline{2-3}
                                   & \multicolumn{1}{c|}{Reliable Sets} & Ambiguous Sets &                                \\ \midrule

No Division                        & \multicolumn{2}{c|}{No Consistency Loss}                        & 50.1                           \\ \midrule

No Division                        & \multicolumn{2}{c|}{KL loss}                        & 56.7                           \\
No Division                        & \multicolumn{2}{c|}{CE loss}                        & 56.1                           \\ \midrule
Confidence-based                   & CE loss ($\mathcal{L}_{r}$)                            & /              & 55.8                           \\
Confidence-based                   & CE loss ($\mathcal{L}_{r}$)                             & KL  loss  ($\mathcal{L}_{a}$)        & 56.5                           \\ \midrule

Ours  w/o   Mix\_Module                             & CE loss ($\mathcal{L}_{r}$)                             & KL loss  ($\mathcal{L}_{a}$)        & 57.6 \\ \midrule
Ours RAC-Net                            & CE loss ($\mathcal{L}_{r}$)                             & KL loss  ($\mathcal{L}_{a}$)        & \textbf{58.4} \\ \bottomrule
\end{tabular}
}
\label{tab:ablation-baseline}
\end{table*}

\begin{figure*}[t]
\centering
    \includegraphics[width=1\linewidth]{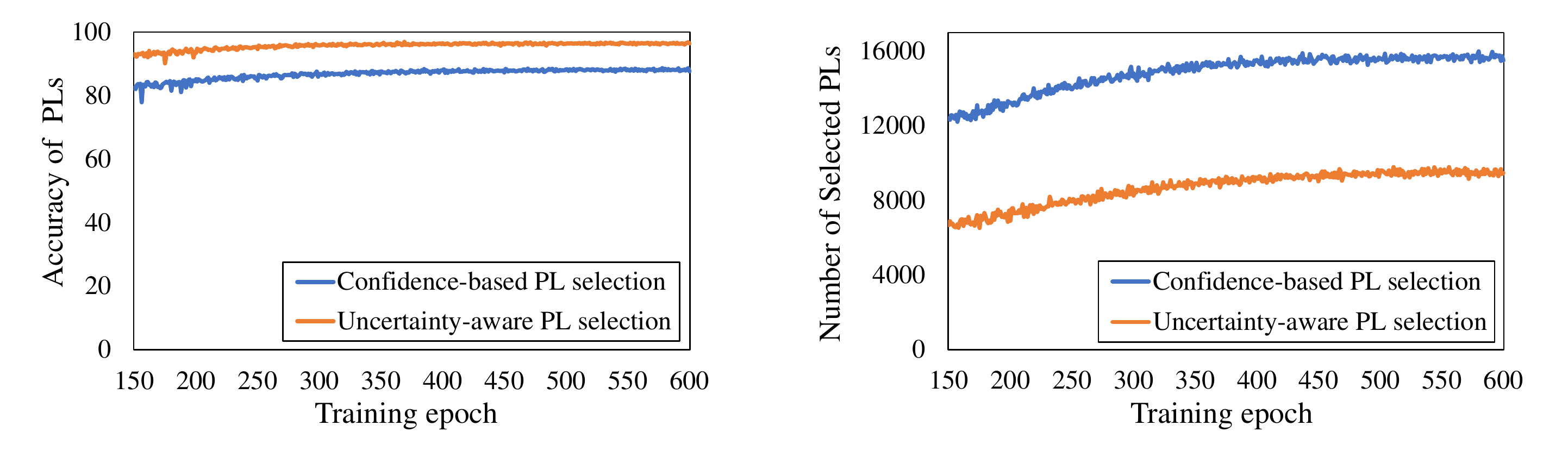}
    \caption{Comparison of the accuracy (left) and the number (right) of selected reliable pseudo labels on the S3DIS training set between the confidence-based and our uncertainty-aware pseudo labeling methods.}
    \label{fig:pl_accuracy}
\end{figure*}

\begin{figure*}[t]
\centering
    \includegraphics[width=0.9\linewidth]{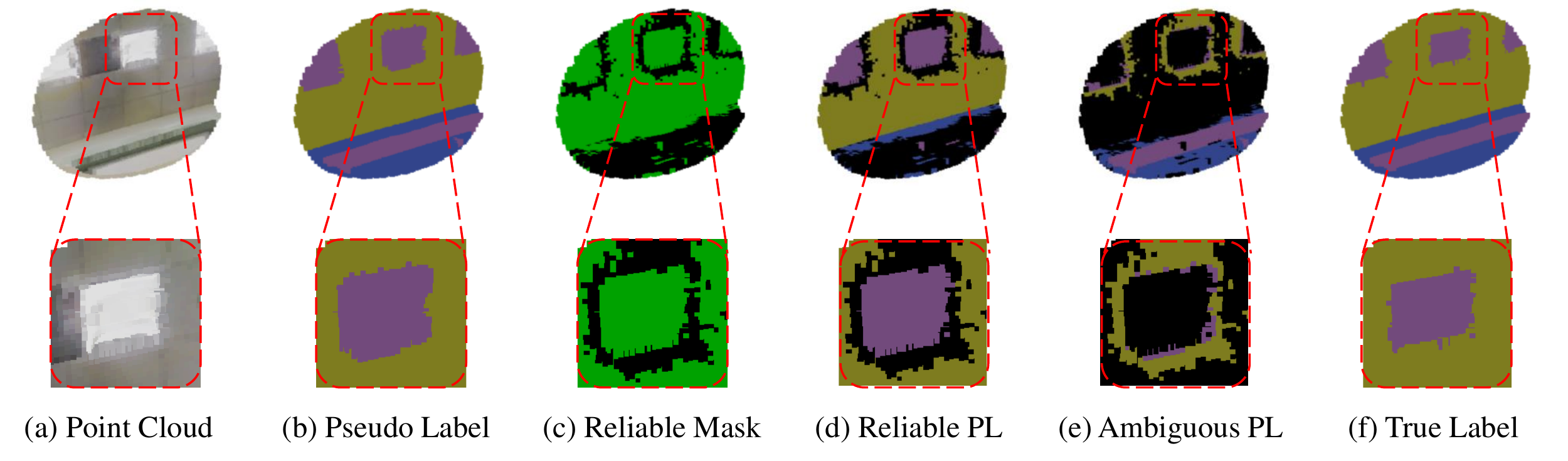}
    \caption {Illustration of the separation of pseudo labels. By jointly considering the uncertainty, our model accurately divides the pseudo labels into the reliable (d) and ambiguous (e) sets. Note that ``black regions'' denotes the masked points.}
    \label{fig:vis-our}
\end{figure*}

\noindent\textbf{Experiment Setting.}
\label{sec:imple}
If there is no special declaration, we implemented our proposed RAC-Net with the KPConv \textit{rigid} backbone, which is identical as the latest DAT model~\citep{wu2022dual}. We used the SGD to train the model with a batch size of 2 and a learning rate of 0.01. We set the confidence threshold $\tau$ as 0.7 and the uncertainty threshold $\kappa$ as 0.05. All experiments in this paper were conducted in an identical environment (Hardware: single NVIDIA RTX 3090 GPU; Software: PyTorch 1.7.0 and CUDA 11.0).

\subsection{Results on the S3DIS dataset}
\subsubsection{Comparing with State-of-the-art Methods}
Table~\ref{tab:s3dis-compare-sota} shows the mIoU results of our proposed RAC-Net and several public methods on the S3DIS Area 5 set. It indicates that our RAC-Net is able to achieve comparable results compared to the fully supervised upper bound, where all points are labeled for the model training. Meanwhile, our model significantly outperforms the latest SOTA method DAT by an 1.9\% mIOU improvement under the ``OTOC'' setting. Under the ``One Thing Three Clicks''(OTTC) setting that three points are labeled for each thing, the proposed model also improves the mIOU by a 2.0\% gain compared with DAT~\citep{wu2022dual}. This demonstrates the effectiveness of our RAC-Net in weakly-supervised point cloud segmentation.

\subsubsection{Discussions}
\noindent\textbf{Ablation studies.} Table~\ref{tab:ablation-loss} gives the ablation studies of our RAC-Net with different losses. We can see that using each loss for training can always improve the segmentation performance, and applying all losses achieves the highest mIoU. It implies that the soft and one-hot pseudo labels are crucial for adaptive consistency regularization and complement each other. Meanwhile, effectively leveraging the reliable pseudo-labels (\textit{i.e.,} using $\mathcal{L}_{mix}$) could further improve the performance.

\noindent\textbf{Comparisons with Baselines.}
As shown in Table~\ref{tab:ablation-baseline}, we constructed three baselines to show the effectiveness of our Separation and Consistency Module.

1) \textit{w/ or w/o division.} The first, second and fifth rows in Table~\ref{tab:ablation-baseline} indicate that the performance with only consistency training via using the KL Divergence or CE loss for all the points is sub-optimal on the S3DIS dataset, and adaptively applying the consistency constraints on ambiguous and reliable sets can achieve better segmentation performance.

2) \textit{w/ or w/o ambiguous sets.} Without considering uncertainty, we use the prediction confidence with a threshold of 0.7 to obtain the reliable and ambiguous sets, as used in the conventional methods \citep{liu2021one}. Then, we adopt a cross-entropy loss to guide the learning of multiple augmented versions on the reliable set. From the third and fourth rows, we can see that further applying the KL Divergence on the ambiguous set for the consistency constraints can improve the mIoU results by 0.7\%, which indicates that effectively exploiting the ambiguous data can facilitate model training.

\begin{table}
\caption{Discussion of three different augmentation methods \textit{i.e.,} PointWolf, Affine Transformation(AT), and Point-wise Random Noise(PRN), under the OTOC setting on the S3DIS dataset.}
\centering
\begin{tabular}{c|ccc|c}
\toprule
\multirow{2}{*}{K} & \multicolumn{3}{c|}{Augmentation Methods}                       & \multirow{2}{*}{mIoU(\%)}      \\ \cline{2-4}
                   & \multicolumn{1}{c|}{PointWolf} & \multicolumn{1}{c|}{AT} & PRN &                                \\ \midrule
\multirow{3}{*}{1} & \checkmark                              &                          &     & 55.8                           \\
                   &                                & \checkmark                        &     & \textbf{57.2}                           \\
                   &                                &                          & \checkmark   & 55.5                           \\ \midrule
\multirow{3}{*}{2} & \checkmark                              & \checkmark                        &     & \textbf{58.4} \\
                   & \checkmark                              &                          & \checkmark   & 56.5                           \\
                   &                                & \checkmark                        & \checkmark   & 56.7                           \\ \midrule
3                  & \checkmark                              & \checkmark                        & \checkmark   & \textbf{58.5} \\ \bottomrule
\end{tabular}
% \vspace{-0.5cm}
\label{tab:ablation-k}
\end{table}

3) \textit{w/ or w/o uncertainty-based division.} Comparing the fourth and the fifth rows of Table~\ref{tab:ablation-baseline}, we can see that employing both confidence and uncertainty for separation can increase the mIoU results by 1.1\%, which suggests that the uncertainty is useful to improve the segmentation performance. Thus, we jointly employ the prediction confidence and uncertainty to divide the unlabeled points in this paper.

In addition, Figure~\ref{fig:pl_accuracy} shows the accuracy and number of selected reliable pseudo labels on the S3DIS training set, of the confidence-based and our uncertainty-aware pseudo-label selection methods. ``Confidence-based PL selection'' indicates that the pseudo labels are selected from the predictions with a confidence threshold 0.7. The left sub-figure of Figure~\ref{fig:pl_accuracy} reveals that, by introducing the uncertainty measurement, the accuracy of the selected pseudo label is able to be improved by around 10\%. On the other hand, the right sub-figure shows the number of selected pseudo labels. We can see that uncertainty-aware pseudo-label selection would have fewer reliable labels. Meanwhile, with the development of model training, the number of selected pseudo labels is increased. Overall, by considering both confidence and uncertainty, our model is able to select fewer but high-quality pseudo labels during the model training.
Besides, we visualize the divided pseudo labels in Figure~\ref{fig:vis-our}. We can see that: (1) most of the pseudo labels in the reliable set are correct; and (2) the ambiguous pseudo labels mainly locate at the blurred and boundary regions, which are indicated by our model.
\begin{table}[t]
\caption{Discussion of using different losses for the model training under the OTOC setting on the S3DIS dataset. Note that, ``R'' denotes the reliable set and ``A'' denotes the ambiguous set.}
\centering
		\begin{minipage}[t]{0.23\textwidth}
		\centering
			\setlength{\tabcolsep}{5pt}
			\begin{center}
				\begin{tabular}{c|c|c}
\toprule
Set                & Type                  & mIoU(\%) \\ \midrule
\multirow{2}{*}{R} & Dice             &  48.8    \\
                     & CE                  &  \textbf{58.4}    \\ \bottomrule
\end{tabular}
			\end{center}
			\begin{center}

			\end{center}
		\end{minipage}
		\begin{minipage}[t]{0.23\textwidth}
		\centering
			\setlength{\tabcolsep}{5pt}
			\begin{center}
				\begin{tabular}{c|c|c}
                \toprule
                Set                & Type                  & mIoU(\%) \\ \midrule
                \multirow{2}{*}{A} & MSE             &  54.1    \\
                     & KL                  &  \textbf{58.4} \\ \bottomrule
                \end{tabular}
			\end{center}
			\begin{center}

			\end{center}
		\end{minipage}
\label{tab:loss_function}
\end{table}

\begin{table}[t]
\caption{Discussion of different $\tau$ and $\kappa$ in Eq.\ref{eq:division} under the OTOC setting on the S3DIS dataset.}
\centering
		\begin{minipage}[t]{0.23\textwidth}
		\centering
			\setlength{\tabcolsep}{5pt}
			\begin{center}
				\begin{tabular}{c|c|c}
\toprule
$\tau$                & $\kappa$                  & mIoU(\%) \\ \midrule
\multirow{3}{*}{0.7} & 0.05                  &  \textbf{58.4}    \\
                     & 0.1                   &  57.7    \\
                     & 0.2                   &  57.6    \\ \bottomrule
\end{tabular}
			\end{center}
			\begin{center}

			\end{center}
		\end{minipage}
		\begin{minipage}[t]{0.23\textwidth}
		\centering
			\setlength{\tabcolsep}{5pt}
			\begin{center}
				\begin{tabular}{c|c|c}
                \toprule
                $\tau$                & $\kappa$                 & mIoU(\%) \\ \midrule
                0.7                  & \multirow{3}{*}{0.05} &  \textbf{58.4}    \\
                0.8                  &                       &  56.9    \\
                0.9                  &                       &  57.0    \\ \bottomrule
                \end{tabular}
			\end{center}
			\begin{center}

			\end{center}
		\end{minipage}
\label{tab:s3dis-ablation-hyper}
\end{table}

\begin{table}[h]
\caption{Discussion of different hyper-parameters under the OTOC setting on the S3DIS dataset.}
\centering
\resizebox{0.5\linewidth}{!}{
\begin{tabular}{ccc|c}
\hline
$\lambda_1$             &$\lambda_2$              & $\lambda_3$             & mIoU (\%) \\ \midrule
0.5                     & 1                       & 1                       & 57.0 \\
1                       & 0.5                     & 1                       & 58.0 \\
1                       & 1                       & 0.5                     & 58.2 \\
1                       & 1                       & 1                       & \textbf{58.4}    \\
\hline
\end{tabular}
}
\label{tab:lamda}
\end{table}

\begin{table}
% \vspace{-1cm}
\caption{Comparison of our RAC-Net model and the DAT model \citep{wu2022dual} on the S3DIS dataset, using MinkowskiNet32 \citep{choy20194d} as the backbone. Here, our model always achieves better performance.}
\centering

\resizebox{\linewidth}{!}{
\begin{tabular}{l|c|c}
\toprule
Method                                                               & Supervision & mIoU (\%)                      \\ \midrule
Baseline (Mink)                                                         & 0.02\% (OTOC)    & 48.7                           \\
DAT (Mink)                                                      & 0.02\% (OTOC)    & 54.6 \\
RAC-Net (Mink)                                                         & 0.02\% (OTOC)    & \textbf{58.6} \\  \midrule
Baseline (Mink)                                                         & 0.06\% (OTTC)    & 55.0                           \\

DAT (Mink)                                                        & 0.06\% (OTTC)    & 58.2 \\

RAC-Net (Mink)                                                        & 0.06\% (OTTC)    & \textbf{59.9} \\ \midrule
Upper Bound (Mink)                                                      & 100\%            & 65.4                           \\ \bottomrule
\end{tabular}
}
\label{tab:s3dis-ablation-mink}
\end{table}

\noindent\textbf{Number of augmentations.}\label{sec:ab_K} Table~\ref{tab:ablation-k} gives the performance of our RAC-Net with all designed modules, regarding different numbers of augmentations $K$.  Here, we explore three different point cloud augmentation methods: PointWolf, Affine Transformation (AT), and Point-wise Random Noise (PRN). \textit{PointWolf} is a deformation method that applies locally weighted transformations to multiple anchor points to achieve smooth non-rigid deformations. \textit{AT} generates the augmentations by applying global-wise rotation, scaling, and translation to the whole point clouds. \textit{PRN} injects small point-wise noises into each point to generate the augmented point clouds. From Table~\ref{tab:ablation-k}, we can see that: (1) all adopted augmentation techniques can improve the mIoU results; (2) jointly using different augmentations can lead to better performance. To balance the efficiency and effectiveness, we finally select $K=2$ with ``PointWolf \& AT'' as the augmentation methods in this paper.

\noindent\textbf{Different loss functions.}
The cross-entropy loss has been widely used for segmentation tasks when the ``groundtruth labels'' are available. Here, since the model predictions in the reliable set are accurate enough (see Fig.~\ref{fig:pl_accuracy}), we then transform the probability predictions into one-hot pseudo labels, which are applied for the model training with the cross-entropy loss.

At the same time, the KL loss is commonly used to enforce consistency constraints between different augmentation with soft pseudo labels. In this paper, instead of neglecting the ambiguous predictions with high uncertainty or low confidence, we consider their probability results as soft pseudo labels and apply the KL loss as a consistency constraint in the ambiguous set.

Table~\ref{tab:loss_function} shows the performance of using different loss functions to train our model on the S3DIS dataset. It reveals that, using the KL loss to apply a consistency constraint while at the same time applying the cross entropy loss for segmentation can achieve the best performance on the S3DIS dataset. 

\noindent\textbf{Different hyper-parameters.}
\label{sec:hyper}
$\tau$ and $\kappa$ are two important hyper-parameters to divide the pseudo labels in our model, which are used in Eq.~\ref{eq:division}. We show the mIoU results of our RAC-Net with different hyper-parameters in Table~\ref{tab:s3dis-ablation-hyper}. It demonstrates that, setting $\tau$ as 0.7 and $\kappa$ as 0.05 can yield the best performance on the S3DIS dataset.

Furthermore, we also conduct a sensitivity experiment to discuss the impacts of loss coefficients $\lambda_1$, $\lambda_2$ and $\lambda_3$. Table~\ref{tab:lamda} shows that setting them all as 1 can achieve the best performance on the S3DIS dataset.

\noindent\textbf{Generalization Ability.} We applied our proposed training strategy to a voxel-based point cloud segmentation framework (MinkowskiNet~\citep{choy20194d}) to verify the generalization ability. The experiments were conducted on the S3DIS dataset under the OTOC and OTTC settings. Table~\ref{tab:s3dis-ablation-mink} shows that the results of our proposed RAC-Net is always better than the baseline and the latest DAT model \citep{wu2022dual} (\textit{i.e.,} 9.9\%/4.0\% and 4.9\%/1.7\% mIoU improvements, respectively), which demonstrates that our method is general and can be easily applied to various frameworks.

\subsection{Results on the ScanNet-v2 dataset}

\begin{table*}
% \vspace{-1.5cm}
\centering
\caption{Comparison of our RAC-Net and several public methods on the ScanNet-v2 test set. ``RAC-Net$\dagger$'' denotes that our model is built upon the 1T1C model~\citep{liu2021one}, as ``DAT$\dagger$'' \citep{wu2022dual}.
}
\resizebox{0.65\linewidth}{!}{
  \begin{tabular}{l|c|c}
    \toprule
    Method & Supervision & mIoU (\%)  \\
    \midrule
    Pointnet++~\citep{qi2017pointnet++} & 100\% &33.9 \\
    PointCNN~\citep{li2018pointcnn} & 100\% & 45.8\\
    MinkowskiNet~\citep{choy20194d} &100\% & 73.6 \\
    Virtual MVFusion~\citep{kundu2020virtual} &100\%+2D & 74.6\\  
    MPRM~\citep{wei2020multi} & subcloud-level & 41.1 \\    
    MPRM+CRF~\citep{wei2020multi} & subcloud-level & 43.2 \\  
    \midrule
    CSC\_LA\_SEM~\citep{hou2021exploring} & 20 points & 53.1 \\
    Viewpoint\_BN\_LA\_AIR~\citep{luo2021pointly}	& 20 points & 54.8 \\
    PointContrast\_LA\_SEM~\citep{xie2020pointcontrast} & 20 points & 55.0 \\
    1T1C~\citep{liu2021one} & 20 points & 59.4\\
    MIL-derived transformer \citep{Yang_2022_CVPR} & 20 points & 54.4\\
    DAT~\citep{wu2022dual} & 20 points & 55.2\\
    DAT$\dagger$~\citep{wu2022dual} & 20 points & 62.3\\
    PointMatch~\citep{wu2022pointmatch}  & 20 points & 62.4\\
    \midrule
    Our Baseline (KPConv) & 20 points  & 51.6\\
    Our RAC-Net (KPConv) & 20 points   &  56.6 \\
    Our RAC-Net$\dagger$ (KPConv) & 20 points   & \textbf{62.6}\\
    Our Upper Bound (KPConv)  & 100\%  & 68.4 \\
    \midrule
    Our Baseline (Point-Transformer) & 20 points  & 56.4\\
    Our RAC-Net (Point-Transformer) & 20 points  & \textbf{63.9}\\
    Our Upper Bound (Point-Transformer)  & 100\%  & 67.9 \\
    \bottomrule
  \end{tabular}
}

\label{tab:scannet-test}
% \vspace{-2cm}
\end{table*}

\begin{table*}[h]
\caption{Discussion of our proposed RAC-Net under the 20 points setting on the ScanNet-v2 validation and test set based on the Point Transformer backbone \cite{zhao2021point}. Note that our method sets the thresholds $\tau$ and $\kappa$ as 0.7 and 0.05, respectively, to employ both prediction confidence and uncertainty to divide the unlabeled points.}
\centering
\resizebox{0.7\linewidth}{!}{
\begin{tabular}{c|cc|c|c}
\toprule
\multirow{2}{*}{Division Strategy} & \multicolumn{2}{c|}{Consistency Loss}               & \multirow{2}{*}{val set mIoU (\%)}  & \multirow{2}{*}{test set mIoU (\%) }   \\ \cline{2-3}
                                   & \multicolumn{1}{c|}{Reliable Sets} & Ambiguous Sets &                                \\ \midrule
No Division                        & \multicolumn{2}{c|}{No Consistency Loss}                        & 58.3  & 56.4                         \\ \midrule
No Division                       & \multicolumn{2}{c|}{KL loss}                        & 59.6   & 58.6                        \\
No Division                       & \multicolumn{2}{c|}{CE loss}                        &  62.0  & 62.7                            \\ \midrule 

Ours w/o   Mix\_Module                                & CE loss ($\mathcal{L}_{r}$)                             & KL loss  ($\mathcal{L}_{a}$)        &    64.2 & 59.9  \\ \midrule 
Our RAC-Net                              & CE loss ($\mathcal{L}_{r}$)                             & KL loss  ($\mathcal{L}_{a}$)        & \textbf{ 67.6   } & \textbf{63.9}\\ \bottomrule
\end{tabular}
}
\label{tab:ablation_scannet}
\end{table*}

Table~\ref{tab:scannet-test} gives the mIoU results on the ScanNet-v2 test set in the “3D Semantic label with Limited Annotations” benchmark. The officially given 20 annotated points are used as the sparse labels to train the model. Here, ``Our Baseline'' indicates that we only use the segmentation loss $\mathcal{L}_{seg}$ to train the model with the limited labels. Using the KPConv as the backbone, our RAC-Net achieves a remarkable performance gain, with a 5.0\% improvement over the baseline. 
Moreover, our RAC-Net is able to be easily combined with the existing weakly-supervised point cloud segmentation methods and can improve their performance. For the variant denoted as ``RAC-Net$\dagger$'' in Table~\ref{tab:scannet-test}, we build our RAC-Net upon the 1T1C model under the 20 points setting on the ScanNet-v2 dataset as \citep{wu2022dual}. Specifically, we first use the 1T1C model to generate the pseudo labels for all the training points. Then we use them to replace the sparse annotations to guide the RAC-Net training. With this training strategy, the performance of 1T1C can be further improved by 3.2\% on the ScanNet-v2 test set. 
Furthermore, in Table~\ref{tab:scannet-test},  we extend our model on the Point Transformer backbone\citep{zhao2021point}, which achieves a 63.9\% mIoU and sets the new state-of-the-art performance on the ScanNet-v2 test set.

\noindent\textbf{Ablation studies.} Table~\ref{tab:ablation_scannet} further gives the ablation studies of our RAC-Net based on the Point Transformer backbone \citep{zhao2021point} on the ScanNet-v2 validation set. We can also see that using each loss for training can significantly improve the segmentation performance, and applying all losses achieves the highest mIoU, which aligns with the observations on the S3DIS dataset. It demonstrates the robustness of our proposed model for different datasets and different backbones for weakly supervised point cloud segmentation.

\begin{table}[h]
\caption{Comparison of our model and public methods under the One Thing One Click (OTOC) setting on the SementicKitti dataset \citep{behley2019semantickitti}.}
\centering
\resizebox{0.9\linewidth}{!}{
\begin{tabular}{c|c|c}
\toprule
Method                                                               & Supervision & mIoU (\%)                      \\ \midrule
SQN \citep{hu2021sqn}                                                        & 0.01\%     & 39.1                           \\
Our RAC-Net                                                         & 0.01\% (OTOC)    & \textbf{45.3}  \\  \midrule
Our Upper Bound                                                      & 100\%            & 62.1                           \\ \bottomrule
\end{tabular}
}
\label{tab:semanticekitti}
\end{table}

\subsection{Results on the SemanticKitti dataset}
Meanwhile, Table~\ref{tab:semanticekitti} shows the segmentation performance of our proposed RAC-Net on the SemanticKitti dataset \citep{behley2019semantickitti}. It reveals that our proposed model outperforms the recent SQN \citep{hu2021sqn} for weakly supervised point cloud segmentation tasks.

\subsection{Qualitative Results}

\begin{figure}
\centering
    \includegraphics[width=\linewidth]{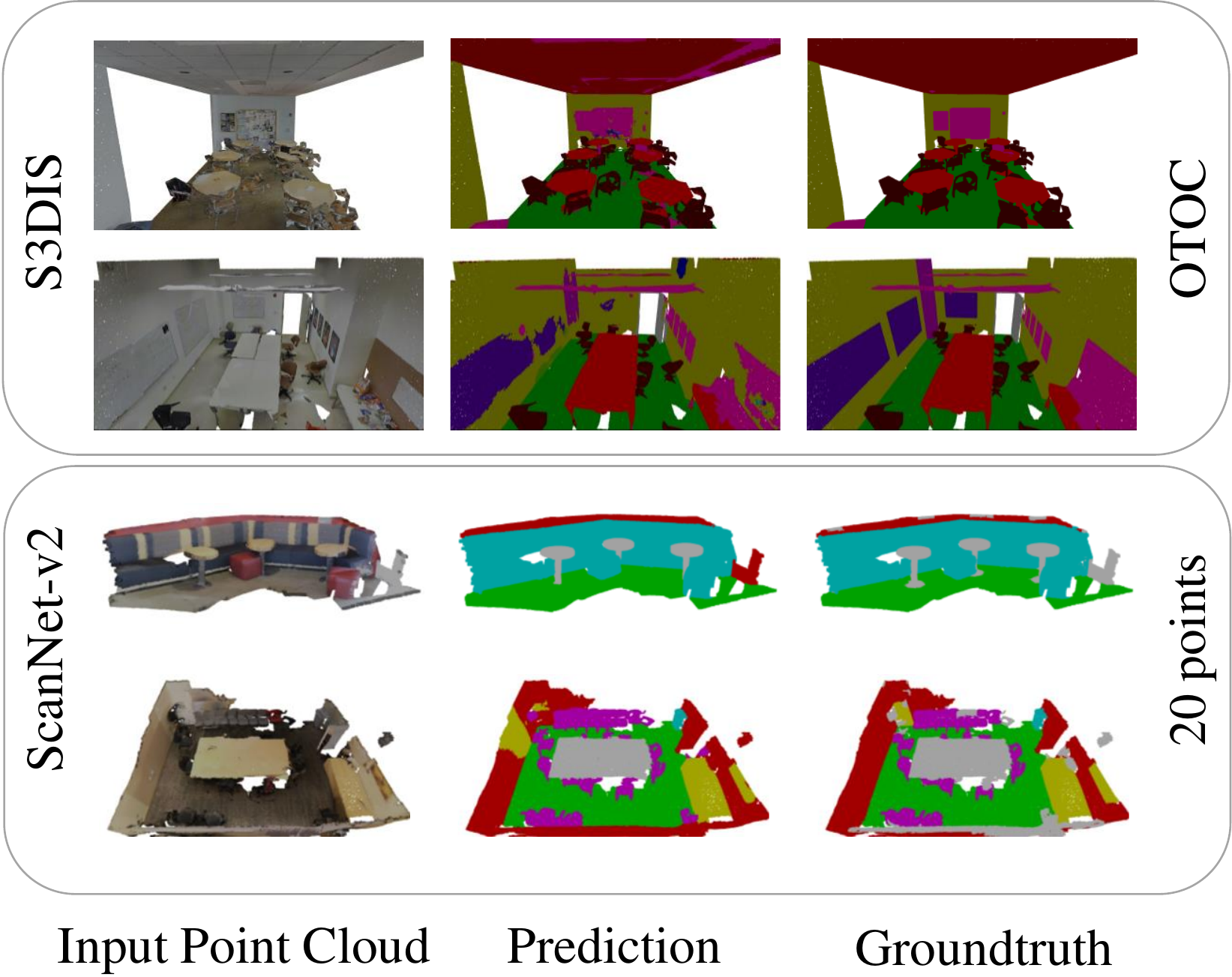}
    \caption{Examples on the S3DIS (top row) and ScanNet-v2 (bottom row) datasets.}
    \label{fig:vis}
\end{figure}

Figure~\ref{fig:vis} shows two exemplar results obtained by our RAC-Net on the S3DIS (under the OTOC setting, top) and ScanNet-v2 (under the 20 points setting, bottom) datasets. With our carefully designed training strategy, the RAC-Net, only trained by sparse annotations and abundant unlabeled points, is able to accurately segment most of the 3D points on both datasets, especially in the small and thin regions. Such an ability is critical for realistic applications. 

\begin{table*}[h]
\caption{Discussion of the total training time, GPU memory, and disk memory usage various methods at different values of K on the S3DIS dataset, using the KPConv backbone.}
\centering
\resizebox{0.7\linewidth}{!}{
\begin{tabular}{c|c|c|c}
\toprule
Method                                             &   Training Time               & GPU memory usage  & Disk memory usage                    \\ \midrule
Only SEG Module (Baseline)                       &  20 hours  & 5 GB & 6.9GB                           \\ 
DAT                        &  120 hours & 9 GB  & 6.9GB                         \\ 
\midrule
Our RAC-Net (K=2)                       &    30 hours                               & 9 GB & 6.9GB \\
Our RAC-Net (K=3)                       &      35 hours                             & 10 GB  & 6.9GB\\\bottomrule  
\end{tabular}
}
\label{tab:ablation-time}
\end{table*}

\subsection{Computational Costs}

During model training, we employ on-the-fly augmentations to each input in an asynchronous manner. Here we do the augmentations and model training simultaneously rather than the model training after the data augmentations for each batch input, resulting in an identical training time with different K values. Regarding the GPU memory consumption, we trained the model on a 2080Ti GPU, consuming roughly 10GB of memory. The usage is only affected by the number of input point clouds associated with the number K. Note that the GPU memory utilization remains unchanged regardless of the augmentation variations. The approximately 6.9GB of disk space is used to store the dataset and the codes. Since augmentations are generated on the fly during model training, they do not require additional disk spaces.

The computation costs of our model using the KPConv backbone on the S3DIS dataset are shown in Table~\ref{tab:ablation-time}.

\subsection{Limitation and Future Work}
\begin{figure}
\centering
    \includegraphics[width=\linewidth]{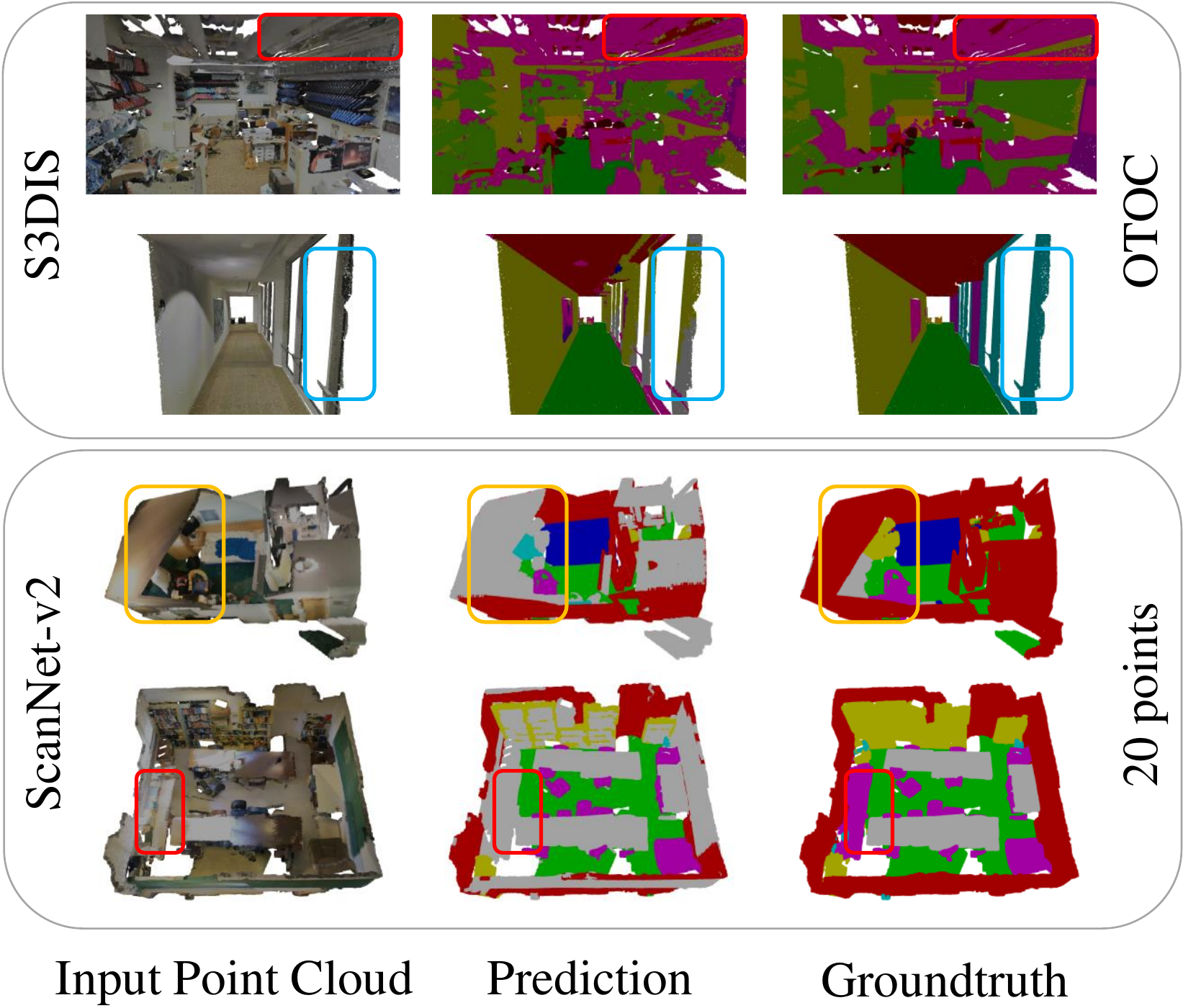}
    \caption{Failure cases for the RAC-Net on the S3DIS (top) and ScanNet-v2 (bottom) datasets. Here, color rectangles indicate different complex scenes (Red: blurred boundaries; Blue: low densities; Yellow: object incompleteness.)}
    \label{fig:failure}
\end{figure}

In general, it is still challenging for current deep models to accurately segment highly complex scenes from the 3D point clouds, especially in the weakly supervised scenario. Figure~\ref{fig:failure} showed several failure cases obtained by our RAC-Net model on the S3DIS and ScanNet-v2 datasets. Specifically, the color rectangles in Figure~\ref{fig:failure} indicate different types of complex scenes such as the blurred boundaries (red), low densities (blue), and object incompleteness (yellow), making it hard to identify individual elements within the scene. Tackling these complex scenes in weakly supervised segmentation will be an interesting future work.

Meanwhile, we utilize the data-specific augmentations in this paper (\textit{e.g.,} PointWolf \citep{kim2021point}, and these techniques are tailored for the point cloud augmentation. Therefore, they cannot be directly applied to other data modalities such as images. However, by incorporating other image-based augmentation methods, our method could be easily generalized to weakly supervised image segmentation.

\section{Conclusion}
\label{sec:conclusion}
In this paper, we have presented a novel RAC-Net to apply the adaptive consistency regularization for weakly-supervised point cloud segmentation. To address the issues of noisy pseudo labels and discarded unreliable points, we propose to jointly use the prediction confidence and uncertainty to select the most accurate pseudo labels and then apply different consistency constraints to different points based on their pseudo label reliability. Via leveraging soft or one-hot pseudo labels on the ambiguous or reliable sets, respectively, our RAC-Net performs the adaptive consistency training to exploit the unlabeled 3D points effectively. Extensive experiments demonstrate that our model outperforms other existing methods and achieves state-of-the-art performance in weakly-supervised point cloud segmentation on both the S3DIS and ScanNet-v2 datasets.

\noindent\textbf{Societal Impacts.} The proposed RAC-Net is trained on two specific point cloud datasets, which would have certain dataset biases and may lead to unconvinced predictions in real applications such as robot navigation.

\section*{Data availability statement}
We conduct experiments on the S3DIS~\citep{armeni20163d}, ScanNet-v2~\citep{dai2017scannet} and SemanticKitti~\citep{behley2019iccv} datasets. The S3DIS dataset can be downloaded from \url{http://buildingparser.stanford.edu/}, the ScanNet-v2 dataset can be downloaded from \url{http://www.scan-net.org/} and the SemanticKitti dataset can be downloaded from \url{http://www.semantic-kitti.org/}.

\section*{Acknowledgement} 
This research is supported by the Agency for Science, Technology and Research (A*STAR) under its MTC Programmatic Funds (Grant No. M23L7b0021).

\bibliography{sn-bibliography}

\end{document}